\PassOptionsToPackage{numbers}{natbib}
\documentclass{article}

\usepackage[preprint]{neurips_2025}
\usepackage{graphicx} 
\usepackage{comment}

\usepackage[utf8]{inputenc} 
\usepackage[T1]{fontenc}    
\usepackage{hyperref}       
\usepackage{url}            
\usepackage{booktabs}       
\usepackage{amsfonts}       
\usepackage{nicefrac}       
\usepackage{microtype}      
\usepackage{xcolor}         
\usepackage{amsmath} 
\usepackage{algorithm}
\usepackage{algpseudocode}
\usepackage{amssymb}
\usepackage{pifont}
\newcommand{\cmark}{\ding{51}}  
\newcommand{\xmark}{\ding{55}}  

\usepackage{pifont,xcolor}

\newcommand{\tick}{\ding{51}}  

\usepackage{bm}

\title{SwarmThinkers: Learning Physically Consistent Atomic KMC Transitions at Scale}

\author{%
  Qi Li\thanks{Work done during an internship at Microsoft Research (in 2024.5),} \\
  University of Science and \\
  Technology of China
  \And
  Kun Li\thanks{Corresponding author: Kun Li (\texttt{likungw@gmail.com})} \\
  Microsoft Research
  \And
  Haozhi Han \\
  Peking University
  \And
  Honghui Shang \\
  University of Science and \\
  Technology of China
  \And
  Xinfu He \\
  Chinese Academy of \\
  Sciences
  \And
  Yunquan Zhang \\
  Chinese Academy of \\
  Sciences
  \AND
  Hong An \\
  University of Science and \\
  Technology of China
  \And
  Ting Cao \\
  Microsoft Research
  \And
  Mao Yang \\
  Microsoft Research
}

\begin{document}

\maketitle

\begin{abstract}
Can a scientific simulation system be physically consistent, interpretable by design, and scalable across regimes—all at once? 
Despite decades of progress, this trifecta remains elusive. Classical methods like Kinetic Monte Carlo ensure thermodynamic accuracy but scale poorly; learning-based methods offer efficiency but often sacrifice physical consistency and interpretability.
We present \textbf{SwarmThinkers}, a reinforcement learning framework that recasts atomic-scale simulation as a physically grounded swarm intelligence system. Each diffusing particle is modeled as a local decision-making agent that selects transitions via a shared policy network trained under thermodynamic constraints.  A reweighting mechanism fuses learned preferences with transition rates, preserving statistical fidelity while enabling interpretable, step-wise decision making. Training follows a centralized-training, decentralized-execution paradigm, allowing the policy to generalize across system sizes, concentrations, and temperatures without retraining. 
On a benchmark simulating Fe–Cu alloy precipitation, SwarmThinkers is the \textit{first} system to achieve full-scale, physically consistent simulation on a single A100 GPU—previously attainable only via OpenKMC on a supercomputer. It delivers up to 4,963× (3,185× on average) faster computation with 485× lower memory usage. By treating particles as decision-makers—not passive samplers—SwarmThinkers marks a paradigm shift in scientific simulation: one that unifies physical consistency, interpretability, and scalability through agent-driven intelligence.

\end{abstract}

\section{Introduction}

Scientific simulation serves as a cornerstone of modern science, enabling insight and prediction across fields—from radiation damage to catalysis to phase transformations~\cite{Domain2019, 10.3389/fchem.2019.00202, gao2018void, Martínez2020, articleDeo, roy2025review}. Among these, Kinetic Monte Carlo~\cite{10.1063/5.0083251, chatterjee2007overview, XU2015135} stands out for modeling rare-event dynamics over extended timescales, grounded in physically derived transition rates~\cite{jansen2003introductionmontecarlosimulations, 10.1063/1.1415500, Donev_2010}.   

However, classical KMC is fundamentally \textbf{passive}:  transitions are sampled blindly from predefined rates, treating all events as equally probable regardless of their downstream impact. In realistic systems, only a small subset of transitions meaningfully shape long-term structure~\cite{10.1063/1.2741391, Allen_2006, P_rez_Espigares_2019, Shannon_2018, papakonstantinou2023hamiltonianmcmcmethodsestimating}. As a result, simulations waste vast compute resolving reversible fluctuations while missing opportunities to prioritize structure-forming transitions. Existing advances such as event cataloging~\cite{NTIOUDIS2023112421, articleTorre, articleMasonDaniel, Oppelstrup_2009}, rejection-free methods~\cite{Mu_oz_2003, yang2010rejectionfreekineticmontecarlo, unknownRuzayqatHamza, articleSlepoyAlexander}, and neural surrogates~\cite{lin2024montecarlophysicsinformedneural, PhysRevB.95.214117} retain this sampling-centric paradigm, often at the cost of scalability, interpretability, or physical consistency.

This raises a deeper question:  \textit{Can particles in simulation learn to think?} That is, can we move beyond passive sampling, toward systems where particles make physically grounded, structure-aware, and goal-directed decisions to drive long-term evolution? 

We propose \textbf{SwarmThinkers}, a new reinforcement learning framework that reimagines atomic-scale simulation as a  \textit{physically constrained swarm intelligence system}.  Each diffusing particle is modeled as an autonomous agent that perceives its local context and selects transitions via a shared policy network.  Rather than sampling independently, all agent-direction pairs are pooled into a unified global softmax, allowing the system to prioritize kinetically meaningful events in a coordinated, structure-aware manner. 
To preserve physical consistency, we introduce a \textit{reweighting mechanism} that fuses policy-driven preferences with classical transition rates, and apply \textit{trajectory-level importance sampling} to ensure unbiased estimation of physical observables and statistics. Physical knowledge is deeply embedded:   action constraints reflect local bonding physics, rewards capture thermodynamic relaxation, and critics encode global structure. Training follows a centralized-training, decentralized-execution (CTDE) paradigm, enabling generalization across system sizes, concentrations, and temperature regimes.

As summarized in Table ~\ref{tab:sota_comparison}, SwarmThinkers uniquely achieves state-of-the-art performance across all metrics: 3,185x faster simulation, $6.68\times10^{-2}$ transition energy per step, and 0.34 effective transition ratio. \textit{For the first time}, a single GPU  scales over 54 billion atoms using less than 60GB of memory, matching the fidelity of previous supercomputer-scale simulations that required over five million cores. From supercomputers to a single GPU, from passive sampling to active reasoning, from scale-bound heuristics to generalizable intelligence—SwarmThinkers represents a paradigm shift toward scalable, physically grounded, agent-driven scientific systems.


\begin{table}[h]
\centering
\caption{
\textbf{Comparison of SwarmThinkers and KMC Acceleration Strategies.}
Metrics include \textbf{Spd} (Speedup over OpenKMC), \textbf{TPE} (Transition Per-step Energy), and \textbf{ETR} (Effective Transition Ratio).
\textbf{Scale} evaluates billion-atom scalability; \textbf{CF} indicates fidelity in cluster formation; 
\textbf{Gen.} assesses generalization across diverse simulation regimes, including different system sizes, temperatures, and Cu/vacancy concentrations.
}
\vspace{0.5em}
\resizebox{\linewidth}{!}{
\begin{tabular}{c|c|c|c|c|c|c|c}
\toprule
\textbf{Cat.} & \textbf{Method} & \textbf{Spd~($\uparrow$)} & \textbf{TPE~(eV, $\uparrow$)} & \textbf{ETR~($\uparrow$)} & \textbf{Scale} & \textbf{CF} & \textbf{Gen.} \\
\midrule
Classic   & OpenKMC~\cite{10.1145/3295500.3356165}              & 1.00x       & $<10^{-5}$  & $<10^{-4}$ & \tick & \tick & \tick   \\
\midrule
ML-Surr.  & MC-PINNs~\cite{lin2024montecarlophysicsinformedneural}             & 1.00x       & $<10^{-5}$  & $<10^{-4}$ & \xmark & \xmark & \cmark \\
ML-Surr.  & NNP-KMC~\cite{PhysRevB.95.214117}              & 1.00x       & $<10^{-5}$  & $<10^{-4}$ & \xmark & \cmark & \xmark \\
\midrule
Biased    & hyper-MD~\cite{voter1997hyperdynamics}             & 15.13x      & $3.19\times10^{-4}$  & $7.81\times10^{-4}$ & \xmark & \cmark & \cmark \\
Biased    & SpecTAD~\cite{10.1177/0037549716674806}              & 15.86x      & $3.25\times10^{-4}$   & $3.17\times10^{-3}$ & \xmark & \cmark & \cmark \\
Biased    & PS-KM~\cite{lin2019partialscaling}                & 62.52x      & $1.32\times10^{-3}$  & $5.34\times10^{-3}$ & \xmark & \cmark & \cmark \\
Biased    & tRPS~\cite{liu2023acceleratingkineticstimereversalpath}                & 77.63x      & $1.65\times10^{-3}$  & $7.29\times10^{-3}$ & \xmark & \cmark & \cmark \\
\midrule
RL-based  & TKS-KMC~\cite{tang2023reinforcementlearningguidedlongtimescalesimulation}    & 1783.18x       & $4.16\times10^{-2}$  & 0.21 & \xmark & \cmark & \xmark \\
RL-based  & PGMC~\cite{Bojesen_2018}              & 2652.54x         & $5.71\times10^{-2}$  & 0.26 & \xmark & \cmark & \xmark \\
\midrule
\textbf{Ours} & \textbf{SwarmThinkers} & \textbf{3185.40x} & $\bm{6.86 \times 10^{-2}}$ & \textbf{0.34} & \textbf{\cmark} & \textbf{\cmark} & \textbf{\cmark} \\
\bottomrule
\end{tabular}
}
\label{tab:sota_comparison}
\end{table}

\section{Related Work}

\subsection{Classical KMC}
Kinetic Monte Carlo models thermally activated processes such as diffusion~\cite{rahman2005atomistic,kratzer2009monte, yang2008kinetic,gillespie2009stochastic}, defect clustering~\cite{articleZhucongand, osti_1666149}, and irradiation damage by sampling transitions from physically derived rate laws~\cite{10.1145/3295500.3356165,becquart2019monte,becquart2012kinetic}.
However, classical KMC remains fundamentally passive: all transitions are sampled solely based on local energy barriers, with no memory, prioritization, or structural feedback. This yields statistically valid but behaviorally unintelligent simulations—unable to focus compute on impactful events or adapt to evolving system dynamics.

\subsection{HPC-Accelerated KMC}
To address computational inefficiency, various acceleration techniques have been proposed. Hyperdynamics~\cite{voter1997hyperdynamics, miron2003parallel, chatterjee2010accurate} and parallel replica methods~\cite{voter1998parrep, perez2009amd, simpson2011parrep} increase event frequency via bias potentials or replica averaging, while adaptive KMC frameworks reduce transition search costs through on-the-fly catalog generation~\cite{henkelman2000neb, xu2008akmc, chatterjee2007overview}. System-level efforts such as OpenKMC~\cite{10.1145/3295500.3356165}, SPPARKS~\cite{mitchell2023spparks}, and TensorKMC~\cite{shang2021tensorkmc} scale to billions of atoms by exploiting massive parallelism. Yet these methods preserve the same core limitation: transition selection remains context-free and rate-based, lacking awareness of system structure or long-term objectives.

\subsection{Learning-Augmented KMC}
Machine learning has been introduced to enhance adaptability and reduce cost. Surrogate models predict transition rates from atomic configurations~\cite{kimari2020application,gonzalez2019machine, lin2024montecarlophysicsinformedneural, PhysRevB.95.214117}, graph neural networks encode structural priors~\cite{reiser2022graphneuralnetworksmaterials, choudhary2021atomistic, fung2021benchmarking}, and reinforcement learning has been used to guide transition selection~\cite{tang2023reinforcementlearningguidedlongtimescalesimulation, Bojesen_2018}. 
Despite these advances, key challenges remain: surrogates often break thermodynamic fidelity, centralized RL fails to generalize to large systems, and most models function as black boxes, limiting scientific interpretability and integration.

\section{Why Scaling Simulation Is Not Enough: Toward Intelligent Physical Modeling}

\subsection{The Brute-Force Ceiling: OpenKMC at Supercomputer Scale}
One of the most critical and computationally demanding challenges in materials simulation is modeling the thermal aging of Fe-Cu alloys in structural steels—a canonical system in materials science~\cite{articleZijunWang, articleCuiSenlin, articleWeiGuojunWang, pssa.201600785}. Over decades of in-service time, nanoscale Cu clusters nucleate and grow, driving embrittlement. Capturing this long-timescale evolution requires simulating billions of atoms over hundreds of years of rare-event dynamics\cite{pssa.201600785, articleCuiSenlin, articleBaiMing, articleYangQigui, articleChristien}.

To date, the only system capable of achieving this is \textbf{OpenKMC}, a massively parallel kinetic Monte Carlo framework that runs on a supercomputer. It successfully reproduces long-timescale Cu precipitation in bcc lattices, with results that match experimental observations in both kinetics and morphology. This marks a milestone: the first atomistic simulation of a full-scale material system with over 100 billion atoms~\cite{10.1145/3295500.3356165}. 

However, this success also exposes the limitations of the underlying paradigm. OpenKMC, like all KMC variants, relies on passive sampling: transitions are selected from rate laws without foresight, feedback, or global coordination. Its scaling stems from raw computational power, not algorithmic intelligence. As systems grow larger and longer in timescale, this brute-force approach encounters a ceiling: \textit{scaling computation alone cannot substitute for scalable intelligence}.

\subsection{What’s Missing: Temporal, Spatial, and Strategic Intelligence}
To break through this ceiling, scientific simulation must evolve beyond passive sampling, toward intelligent, structure-aware reasoning. We identify three core capabilities that current simulation paradigms lack:

\textbf{Temporal reasoning.} The ability to plan across steps—to foresee multi-hop pathways or delayed rewards, and avoid shortsighted decisions that trap systems in suboptimal states.

\textbf{Spatial awareness.} The ability to act in context—understanding defect fields, long-range correlations, and global structure when choosing transitions, rather than sampling myopically.

\textbf{Strategic adaptivity.} The ability to learn from failure—suppressing unproductive transitions, reusing successful pathways, and dynamically allocating attention where change matters most.

\begin{figure}[htbp]
    \centering
    \includegraphics[width=\textwidth]{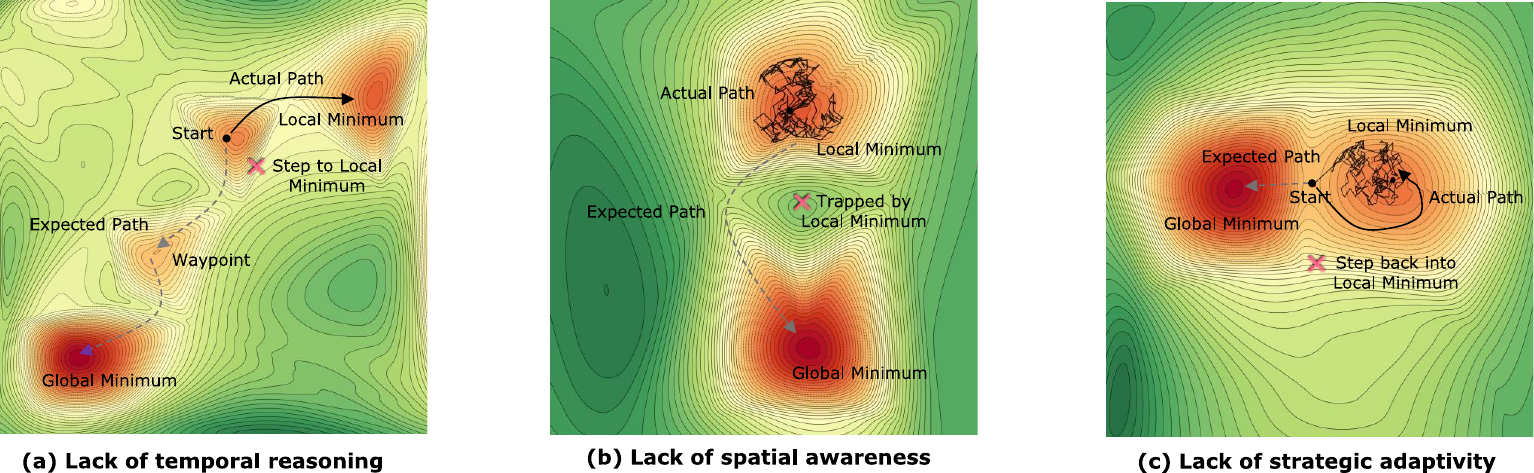}
    \caption{
        Illustration of key limitations in conventional KMC paradigms.
        (a) Lack of temporal reasoning: short-sighted transitions lead to early trapping in local minima, missing delayed but critical structural pathways. 
        (b) Lack of spatial awareness: local decisions ignore global structure, causing the system to wander within energetically unfavorable basins. 
        (c) Lack of strategic adaptivity: Even when near the global minimum, the system may revert to previously explored local minima, failing to suppress historically unproductive transitions.
    }

    \label{fig:drawback}
\end{figure}

These capabilities remain absent in existing paradigms, which is illustrated in Figure~\ref{fig:drawback}. Passive sampling frameworks, even when massively scaled, treat every event in isolation. What’s needed is a simulation system that integrates \textit{local decision-making with global physical constraints}—where particles no longer merely react, but begin to act with purpose.

\section{Method}
\subsection{Preliminary: Kinetic Monte Carlo Method}


To ground our framework, we first review the classical Kinetic Monte Carlo formulation widely used for simulating long-timescale dynamics in crystalline solids. In body-centered cubic (bcc) systems such as Fe–Cu alloys, diffusion proceeds via vacancy-mediated hops, where a vacancy exchanges positions with one of its neighboring atoms. The transition rate of a vacancy-atom hop event \( X \) follows the Arrhenius form:
\begin{equation}
\Gamma_X = \Gamma_0 \exp\left( -\frac{E_a^X}{kT} \right), \quad
E_a^X = E_a^0 + \tfrac{1}{2}(E_f - E_i),
\label{eq:rate_combined}
\end{equation}
where \( \Gamma_0 \) is the attempt frequency, \( k \) the Boltzmann constant, and \( T \) the temperature. \( E_a^0 \) is a species-specific reference barrier, and \( E_i, E_f \) are system energies before and after the hop. These energies are evaluated using a chemically resolved pairwise potential:
\begin{equation}
E_{\text{pair}} = \sum_i n_i \varepsilon^{(i)}_{\text{type}},
\label{eq:pair}
\end{equation}
where \( n_i \) is the number of bonds of type \(\text{type}\) in the \( i \)-th neighbor shell, and \( \varepsilon^{(i)}_{\text{type}} \) the corresponding bond energy. This local model captures short-range chemical effects in dilute alloys.

Given the full set of transition rates \( \{\Gamma_X\} \), the system evolves using the residence-time algorithm. At each step, an event is selected according to: 
\begin{equation}
\Gamma_{\text{tot}} = \sum_X \Gamma_X, \quad
\mathbb{P}(X) = \frac{\Gamma_X}{\Gamma_{\text{tot}}},
\label{eq:total_prob_combined}
\end{equation}
and the time increment is sampled as:
\begin{equation}
\Delta t = -\frac{\ln r}{\Gamma_{\text{tot}}}, \quad r \sim \mathcal{U}(0,1),
\label{eq:residence}
\end{equation}
where $r$ is re-sampled at each KMC step to reflect stochastic evolution. 

\subsection{Particles That Think: Learning to Prioritize Transitions}
\label{sec:reinforcement}
We design a reinforcement learning framework with four key components, which enable atomic-scale agents to prioritize kinetically significant transitions while preserving alignment with first-principles physics.

\textbf{Local Observation.}  
Each agent \( i \) encodes its atomic environment by observing a fixed-radius neighborhood \( \mathcal{N}_i \) consisting of first- and second-nearest neighbors. This domain-informed locality reflects the short-range nature of interatomic interactions, where transition barriers are determined primarily by discrete species configurations rather than geometric coordinates.

We represent the neighborhood as a fixed-length vector:
\[
o_i = \left[\, \sigma_{ij} \,\right]_{j \in \mathcal{N}_i}, \quad o_i \in \mathbb{R}^{n},
\]
where \( \sigma_{ij} \in \mathbb{Z} \) denotes the atomic species of neighbor \( j \), and \( n = |\mathcal{N}_i| \) is the total number of neighbors. This minimalist, symmetry-aware encoding provides a lightweight yet physically grounded state representation.

Importantly, this discrete, topology-preserving encoding captures all transition-relevant factors while abstracting away global details, yielding an input space invariant to system size, defect density, and geometry. It thus provides a scalable foundation for structure-aware decision-making across domains.

\textbf{Decentralized Policy.}  
To reflect the global event selection nature of KMC, we adopt a decentralized policy where each diffusing agent proposes local transitions that are jointly prioritized at the system level. This design decouples local representation from global selection: each agent encodes its neighborhood \( o_i \in \mathbb{R}^d \) via an MLP to produce directional logits: 
\begin{equation}
z_i = f_\theta(o_i), \quad z_i \in \mathbb{R}^K,
\label{eq:policy_logits}
\end{equation}
The final decision emerges from a global coordination process. Specifically, all agent-direction logits are flattened into a single vector:
\begin{equation}
z = \mathrm{concat}(z_1, z_2, \dots, z_N), \quad z \in \mathbb{R}^{NK},
\label{eq:global_logits}
\end{equation}
and passed through a softmax to yield a joint distribution over the entire candidate set:
\begin{equation}
\pi_\theta(a \mid o_{1:N}) = \frac{\exp(z_a)}{\sum_{a'=1}^{NK} \exp(z_{a'})}, \quad a \in \{1, \dots, NK\}.
\label{eq:global_policy}
\end{equation}


The global softmax serves as a differentiable arbitration layer, enabling direct competition among all agent-proposed transitions while retaining localized representations. This fosters swarm-level structural awareness, allowing globally significant transitions to dominate even among locally similar candidates. Its differentiability supports end-to-end training guided by long-horizon structural objectives. Crucially, the learned policy \( \pi_\theta \) is not executed directly, but modulated by physical rates to preserve thermodynamic consistency (see Sec .~\ref{subsec:thermo}). This separation of intent and execution enables intelligent yet physically valid sampling.




\textbf{Centralized Critic.}  
To complement the local policy, we introduce a centralized value function \( V_\phi(s) \) for system-level credit assignment. The input is constructed as \( s = \mathrm{concat}(s_{\text{obs}}, s_{\text{stat}}) \), where
\begin{equation}
s_{\text{obs}} = \mathrm{concat}(o_1, \dots, o_N), \quad
s_{\text{stat}} = \mathrm{concat}(\boldsymbol{\mu}, \boldsymbol{\sigma}, c, d, \rho).
\label{eq:critic_structure}
\end{equation}

Here, \( \boldsymbol{\mu}, \boldsymbol{\sigma} \in \mathbb{R}^3 \) are the spatial mean and variance of Cu positions; \( c \), \( d \), and \( \rho \in \mathbb{R} \) represent local Cu density heterogeneity, average vacancy–Cu distance, and Cu clustering around vacancies, respectively. These descriptors capture key mesoscopic features linked to phase separation, aggregation, and long-range defect interactions. By fusing microscopic observations with coarse-grained statistics, the critic enables scale-bridging credit assignment, linking short-term transitions to long-term structural outcomes. The value function is implemented as a multilayer perceptron:
\begin{equation}
v = f_\phi(s), \quad v \in \mathbb{R},
\label{eq:critic_output}
\end{equation}
and produces a scalar used to compute advantage estimates during policy optimization.


This globally informed signal enhances the policy's ability to discern kinetically meaningful transitions, reduces variance in training across heterogeneous spatial configurations, and stabilizes credit assignment in large-scale systems.

\textbf{Reward and Training.}  
We adopt a CTDE framework for scalable, structure-aware learning. Agents receive only local observations \( o_i \), while the critic leverages global context to assign long-range credit, enabling localized execution with system-level feedback. The reward is defined as the negative change in total system energy:
\begin{equation}
r_t = -\Delta E_t, \quad \Delta E_t = E_{t+1} - E_t,
\label{eq:reward}
\end{equation}
where \( E_t \) denotes the system energy at time step \( t \). This \textit{physics-grounded minimalist reward} reflects the thermodynamic drive toward lower-energy states, guiding the policy to discover transition sequences that accelerate structural relaxation. Crucially, we avoid task-specific shaping, preserving fidelity to the underlying physics and allowing the system to infer meaningful dynamics from first principles.

We optimize the policy using proximal policy optimization (PPO), with advantage targets derived from the centralized critic. By combining local proposals via a global softmax and evaluating outcomes through structure-aware value estimates, the framework enables \textit{delayed credit assignment}—linking long-term morphological changes to earlier local transitions.

This architecture enables strong generalization: since agents rely solely on local observations and actions, the learned policy is invariant to system size, geometry, and defect morphology. Models trained on small domains can thus be deployed directly to large, heterogeneous systems, supporting fast, physically consistent simulation across scales.

\subsection{Embedding Thermodynamics: Preserving Statistical Consistency}
\label{subsec:thermo}

While the learned policy \( \pi_\theta(a \mid o_{1:N}) \) encodes structural preferences that accelerate transitions, physical fidelity must be preserved. Classical KMC transitions are sampled strictly according to Eq.~\ref{eq:total_prob_combined}. Replacing this with a purely learned distribution would violate real-time dynamics.

\textbf{Reweighted Sampling Formulation.}  
To reconcile structure-aware prioritization with thermodynamic correctness, we construct a reweighted distribution over all agent-direction pairs \( a \in \mathcal{A} \):
\begin{equation}
P(a) = \frac{\pi_\theta(a \mid o_{1:N}) \cdot \Gamma_a}{\sum_{a'} \pi_\theta(a' \mid o_{1:N}) \cdot \Gamma_{a'}},
\label{eq:reweighted_prob}
\end{equation}
The hybrid \( P(a) \propto \pi_\theta(a) \Gamma_a \) enhances sampling efficiency but deviates from the physical distribution. We correct this bias via \textit{Importance Sampling} to ensure statistical consistency.

\textbf{Unbiased Estimation.}  
Let \( p(a) \) denote the original KMC distribution and \( q(a) \) the policy-modulated proposal:
\begin{equation}
p(a) = \frac{\Gamma_a}{\sum_{a'} \Gamma_{a'}}, \quad
q(a) = \frac{\pi_\theta(a) \Gamma_a}{\sum_{a'} \pi_\theta(a') \Gamma_{a'}},
\label{eq:pq_def}
\end{equation}
Given samples from \( q(a) \), we compute unbiased estimates of physical observables via:
\begin{equation}
\mathbb{E}_p[f(a)] = \sum_a f(a) \cdot \frac{p(a)}{q(a)} \cdot q(a)
= \frac{Z'}{Z} \sum_a \frac{f(a)}{\pi_\theta(a)},
\label{eq:unbiased_exp}
\end{equation}
with \( Z = \sum_a \Gamma_a \), \( Z' = \sum_a \pi_\theta(a) \Gamma_a \). Since \( \Gamma_a \) cancels due to shared support, the estimator depends only on the inverse policy weight, leading to a practical self-normalized estimator:
\begin{equation}
\mathbb{E}_p[f(a)] \approx \sum_{i=1}^M f(a^{(i)}) \cdot \frac{1 / \pi_\theta(a^{(i)})}{\sum_{j=1}^M 1 / \pi_\theta(a^{(j)})}, \quad a^{(i)} \sim q(a).
\label{eq:unbiased_estimator}
\end{equation}
This guarantees unbiased estimates as long as \( \pi_\theta(a) > 0 \) for all transitions.

\textbf{Trajectory-Level Correction.}  
Beyond step-wise observables, we extend this framework to trajectory-level estimation. For a transition path \( \tau = (a_1, \dots, a_T) \), the cumulative importance weight is:
\begin{equation}
w(\tau) = \prod_{t=1}^{T} \frac{p(a_t)}{q(a_t)} = \left( \frac{Z'}{Z} \right)^T \prod_{t=1}^{T} \frac{1}{\pi_\theta(a_t)}.
\label{eq:trajectory_weight}
\end{equation}
This enables unbiased estimation of path-dependent observables—e.g., advancement factor of Cu. Although variance grows with trajectory length, it can be mitigated via entropy regularization or bounded-horizon estimation.

In summary, this formulation enables structure-aware acceleration of rare-event sampling while preserving statistical and thermodynamic consistency across both steps and trajectories—a principled fusion of learned intent and physical rigor.

\section{Experiments}
\label{sec:experiments}

We evaluate the proposed method from five perspectives: correctness validation, sampling efficiency, large-scale scalability, physical visualization, and comparison with state-of-the-art baselines.

\subsection{Correctness Verification}

To assess physical fidelity, we evaluate the advancement factor \( \zeta(t) \), which measures configurational evolution over time. Results are compared against OpenKMC with pair potentials and experimental data from Lê et al.~\cite{le1992precipitation} and Vincent et al.~\cite{vincent2006solute} at 663–773\,K.

\begin{figure}[htbp]
    \centering
    \includegraphics[width=\textwidth]{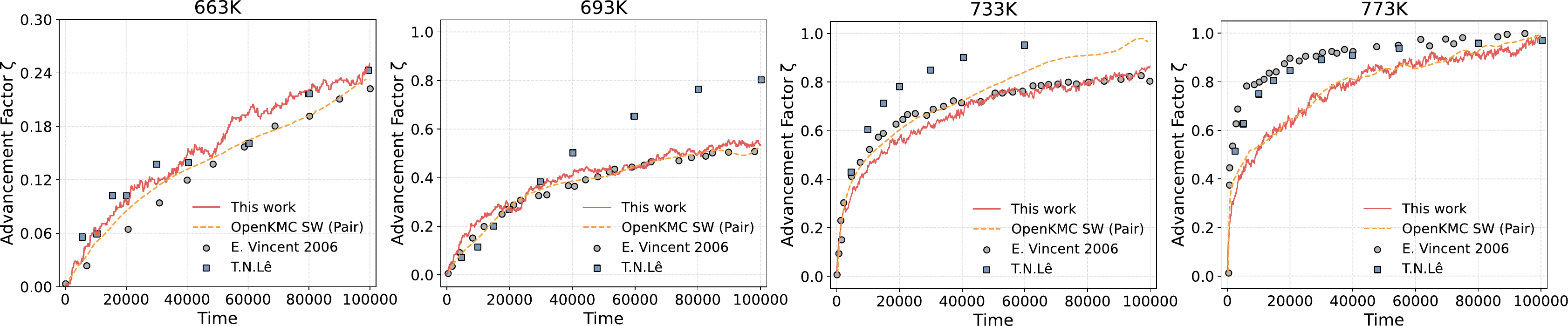}
    \caption{
        \textbf{Advancement factor $\zeta(t)$ versus time under four thermal conditions (663\,K to 773\,K).} 
        Our method (red solid line) shows close agreement with OpenKMC using short-range pair potentials (orange dashed) and aligns well with experimental benchmarks from Vincent et al.~\cite{vincent2006solute} (black circles) and Lê et al.~\cite{le1992precipitation} (gray squares). 
        The results demonstrate the physical correctness of the learned policy across a broad activation spectrum.
    }
    \label{fig:correctness}
\end{figure}

As shown in Figure~\ref{fig:correctness}, our method yields advancement curves that closely follow both simulation baselines and experimental measurements. At lower temperatures (e.g., 663\,K), where transitions are dominated by high barriers, our model reproduces the slower configurational evolution observed experimentally. At higher temperatures (e.g., 773\,K), our approach captures the rapid saturation behavior with high accuracy. These results confirm that the proposed model preserves the thermodynamic pathways and kinetics characteristic of atomic-scale diffusion processes.

\subsection{Sampling Efficiency}

We evaluate sampling efficiency using the \emph{Speedup Ratio}, which quantifies how many more steps a conventional KMC algorithm must take to achieve the same structural evolution as our RL-augmented method. A higher ratio indicates more efficient progression through kinetically meaningful transitions.

Across all Cu alloy systems, our method delivers substantial acceleration over conventional KMC, with the Speedup Ratio peaking at 9361x (4.02\,at.\% Cu, 663\,K). Even in thermally activated, defect-rich regimes (5.36\,at.\% Cu, 773\,K), it sustains over 850x speedup on average. In dilute conditions (1.34\,at.\% Cu), median gains remain above 1500x at 663\,K and 500x at 773\,K. These results highlight the method’s consistent efficiency and broad applicability across diverse compositional and kinetic landscapes.

\begin{figure}[htbp]
    \centering
    \includegraphics[width=\textwidth]{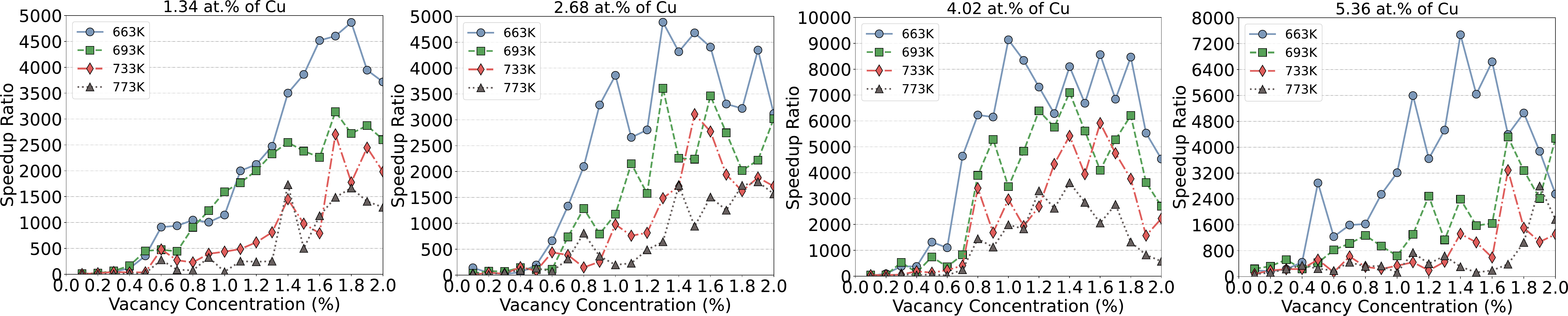}
    \caption{
        \textbf{Sampling acceleration across different Cu alloy systems.} Speedup Ratio is defined as the number of KMC steps required to match the structural evolution achieved by RL-based sampling. Each panel shows performance under varying vacancy concentrations for a fixed Cu composition (1.34--5.36 at.\% Cu). Higher ratios reflect more efficient exploration of the configuration space. Our method maintains robust acceleration across temperature and composition regimes.
    }
    \label{fig:speedup}
\end{figure}

\subsection{Supercomputing-Scale Simulation}

\begin{figure}[htbp]
    \centering
    \includegraphics[width=\textwidth]{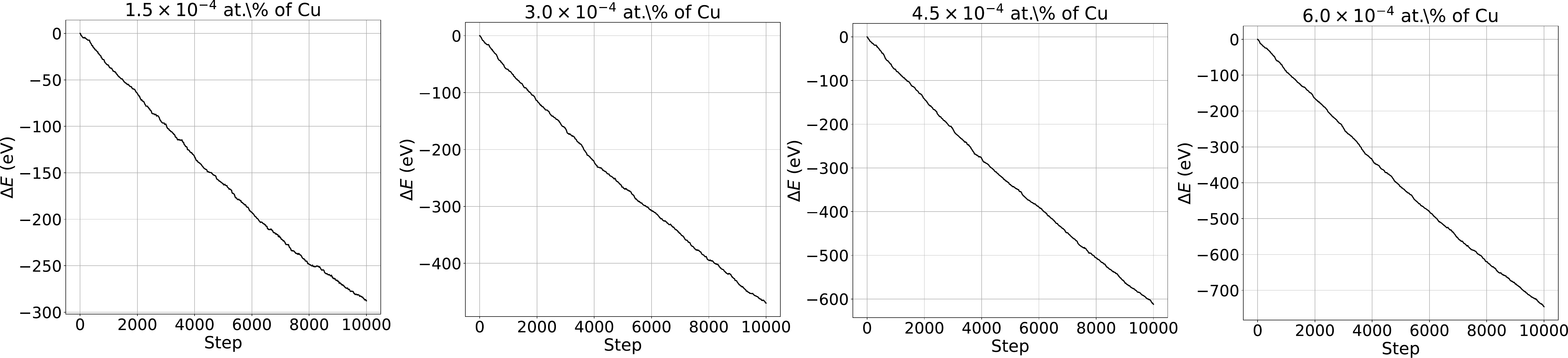}
    \caption{
        \textbf{Supercomputing-scale relaxation on a single GPU.} 
        Each panel plots the cumulative system energy change \( \Delta E \) over 10,000 KMC steps for increasingly concentrated Cu alloys (1.5–6.0 × 10\(^{-4}\) at.\%).
    }
    \label{fig:energy_trend}
\end{figure}
\label{sec:super}
We demonstrate the scalability of our method by simulating Cu alloy systems containing up to 54 billion (5.4 × 10\(^ {10}\)) atoms, matching the problem scales previously handled only by massively parallel supercomputers such as Sunway TaihuLight. In contrast, our method executes them on a single NVIDIA A100 GPU.

Figure~\ref{fig:energy_trend} shows the cumulative energy decrease \( \Delta E \) across 10,000 steps under four Cu compositions (1.5–6.0 × 10\(^{-4}\) at.\%). All systems exhibit smooth, monotonic relaxation, confirming the physical robustness of the learned policy under high-defect, large-scale regimes. No instability or divergence is observed, even under extreme concentrations, validating both the accuracy of the kinetic modeling and the generalizability of the policy.

\begin{figure}[htbp]
    \centering
    \includegraphics[width=\textwidth]{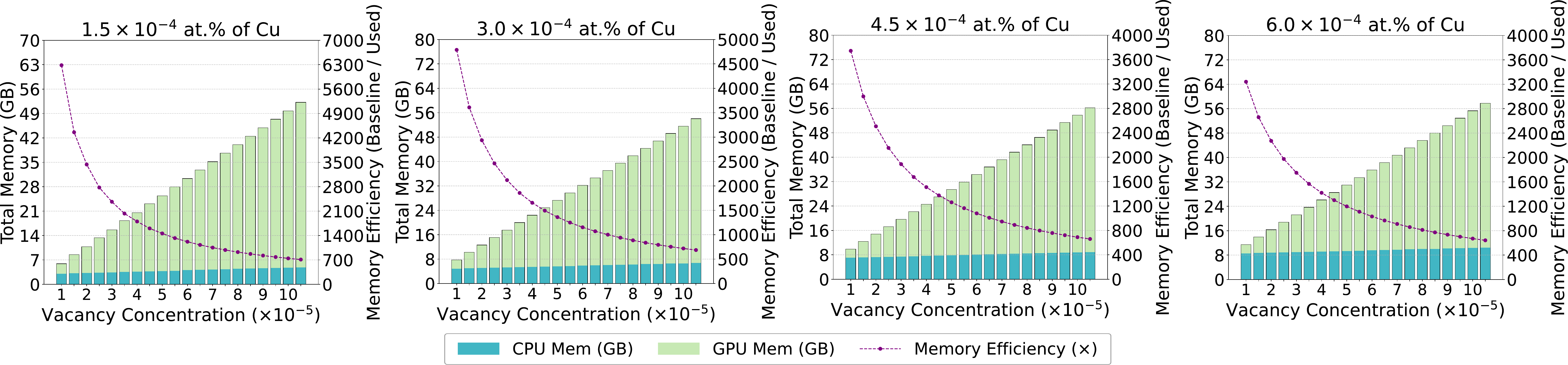}
    \caption{
        \textbf{Memory efficiency comparison with OpenKMC.} 
        Each panel shows the total memory usage (CPU + GPU) across vacancy concentrations for a fixed Cu composition. Bars indicate actual peak usage of our method, while the purple dashed line denotes \textit{Memory Efficiency}—the ratio of OpenKMC's estimated baseline memory on Sunway to ours.
    }
    \label{fig:memory_efficiency}
\end{figure}

To quantify resource efficiency, we compare peak memory usage against OpenKMC—the state-of-the-art conventional KMC implementation—whose baseline consumption exceeds 38.8 TB for comparable systems. As shown in Figure~\ref{fig:memory_efficiency}, our method consistently maintains total memory below 60GB, achieving memory efficiency gains exceeding 500×. This reduction is made through adaptive GPU-CPU memory management and redundant transition avoidance.

These results demonstrate that our system enables physically faithful simulation of previously supercomputer-only workloads using a single commodity GPU, offering a practical and efficient path to long-timescale, high-fidelity KMC at unprecedented scale.

\subsection{Physical Visualization}

\begin{figure}[htbp]
    \centering
    \includegraphics[width=\textwidth]{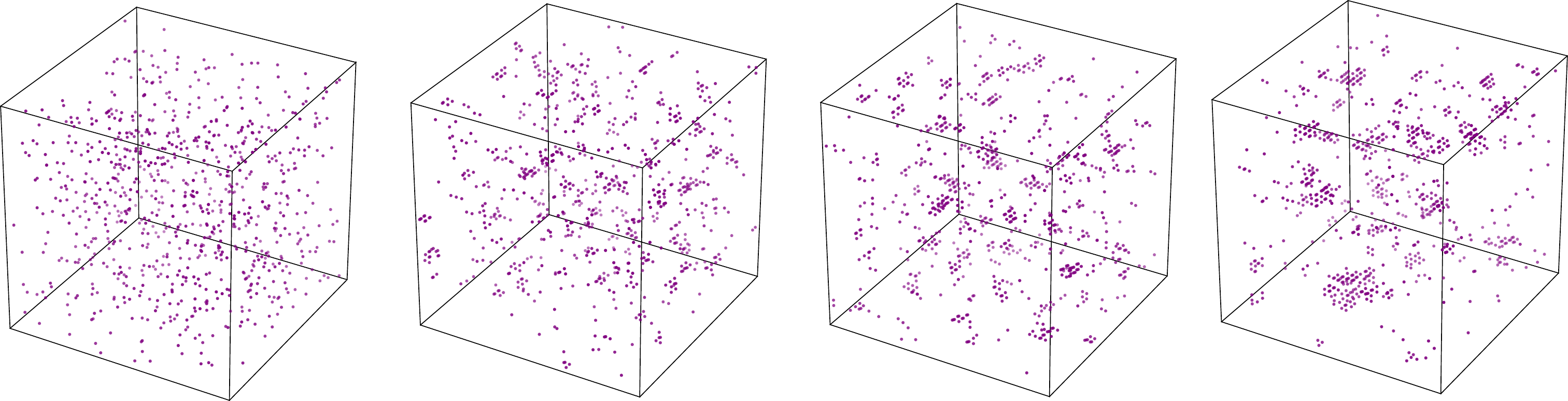}
    \caption{
        \textbf{Cu clustering evolution in Fe–0.67 at.\% Cu over 50 years.} 
        Snapshots show representative configurations at four physical time points simulated under SwarmThinkers. Starting from an initially dispersed distribution, Cu atoms progressively form and coarsen into nanoscale clusters, faithfully reproducing the thermally activated phase separation process observed in RPV steels.
    }
    \label{fig:visualization}
\end{figure}

Figure~\ref{fig:visualization} illustrates the 50-year evolution of Cu atoms in Fe–0.67 at.\% Cu, a canonical aging scenario in structural steels. The system undergoes a two-stage transformation: initial nucleation of clusters followed by long-term coarsening, matching experimentally observed phase separation behavior.

This trajectory highlights SwarmThinkers’ ability to autonomously uncover physically valid pathways under dilute, high-barrier conditions, without relying on handcrafted rules or transition catalogs. Unlike conventional KMC, which often becomes trapped in metastable states, our method consistently drives the system toward thermodynamically favorable configurations, offering both predictive accuracy and interpretable structural insight.

\section{Limitations and Future Work}
\label{sec:limitations}
This work introduces a scalable learning paradigm for rare-event atomistic simulation, achieving high-fidelity evolution in billion-atom systems on a single GPU. While we focus on binary alloys and single-node deployment, the underlying framework is broadly extensible toward more complex chemistries and larger computational scales.

\textbf{Multi-component generalization.} Our current implementation is limited to binary Cu systems. However, the agent-centric architecture is inherently species-agnostic and topology-aware, allowing seamless extension to chemically disordered or high-entropy alloys. This enables direct simulation of complex real-world materials—from steels to aerospace-grade superalloys—without relying on surrogate approximations or coarse-grained abstractions.

\textbf{Distributed scalability.} We have not yet implemented multi-GPU or multi-node execution. Nonetheless, the communication-free nature of agent rollouts makes the framework intrinsically parallelizable. With distributed deployment, SwarmThinkers could exceed the atomistic scale of previous supercomputer-based efforts, advancing toward trillion-atom, long-timescale simulations that resolve realistic microstructural evolution in full physical fidelity.

\section{Conclusion}
We introduced SwarmThinkers, a reinforcement learning framework that redefines atomic-scale simulation as a physically grounded, agent-driven decision process. By unifying physical consistency, interpretability, and scalability, SwarmThinkers enables structure-aware simulations that reason, rather than sample. It is the first system to conduct full-scale precipitations scaling to 54 billion atoms on a single GPU with less than 60GB of memory, previously possible only with supercomputers. 

\newpage

\appendix

\section{Experimental Details}
\label{sec:exp_details}

Table~\ref{tab:exp_config_summary} summarizes the key configurations used across the four primary evaluation tasks presented in this work: correctness verification, sampling efficiency assessment, large-scale scalability testing, and physical trajectory visualization. These experiments are designed to comprehensively evaluate the accuracy, efficiency, and scalability of the proposed RL-augmented KMC framework.

In the \textbf{Correctness} task, we simulate Fe–1.34 at.\% Cu systems at four temperatures (663–773\,K) to validate that our model reproduces thermodynamically consistent kinetics, using advancement factor $\zeta(t)$ as a diagnostic metric. The \textbf{Sampling Efficiency} task spans a broader range of Cu compositions (1.34–5.36 at.\%) under the same temperature settings to quantify acceleration over conventional KMC.

The \textbf{Scalability} task focuses on ultra-large-scale simulations with up to 54 billion atoms, evaluating both physical stability and memory performance. Here, we simulate systems with extremely dilute Cu concentrations (0.015–0.06 at.\%) in a $2000 \times 3000 \times 9000$ lattice. Finally, the \textbf{Visualization} task models Cu clustering over 50 years of physical time at Fe–0.67 at.\% Cu, providing interpretable atomistic trajectories that align with experimental observations in RPV steels.

All experiments share a consistent reinforcement learning architecture. Both the actor and critic networks use a 5-layer multilayer perceptron (MLP) with 256 hidden units per layer and ReLU activations. The system size remains fixed at $40 \times 40 \times 40$ for most experiments, except for the scalability benchmark. Each experiment adopts physically meaningful temperature and composition settings to stress-test the method under varying diffusion regimes.

\begin{table}[h]
\centering
\caption{Summary of experimental settings across the four core evaluation tasks, including composition, temperature, system size, and RL model configurations. Each column corresponds to a distinct experiment.}
\resizebox{\textwidth}{!}{
\begin{tabular}{l|c|c|c|c}
\toprule
\textbf{Aspect} & \textbf{Correctness} & \textbf{Sampling Efficiency} & \textbf{Scalability} & \textbf{Visualization} \\
\midrule
\textbf{Cu composition} & 1.34 at.\% & 1.34 / 2.68 / 4.02 / 5.36 at.\% & 1.5 / 3.0 / 4.5 / 6.0 $\times 10 ^ {-4}$ at.\% & 0.67 at.\% \\
\midrule
\textbf{Temperature (K)} & \multicolumn{2}{c|}{663 / 693 / 733 / 773} & 663 & 663 \\
\midrule
\textbf{System scale} & \multicolumn{2}{c|}{$40\times40\times40$} & $2000\times3000\times9000$ & $40\times40\times40$ \\
\midrule
\textbf{Actor network} & \multicolumn{4}{c}{5-layer MLP: [256, 256, 256, 256, 256], ReLU} \\
\midrule
\textbf{Critic network} & \multicolumn{4}{c}{5-layer MLP: [256, 256, 256, 256, 256], ReLU} \\
\midrule
\textbf{Purpose} & Validate $\zeta(t)$ & Measure Speedup Ratio & Memory and compute scaling & Cu clustering over 50 years \\
\bottomrule
\end{tabular}
}
\label{tab:exp_config_summary}
\end{table}

\begin{table}[h]
\centering
\caption{Summary of the reinforcement learning training configuration used consistently across all experiments. Parameters are grouped into two categories: RL strategy (e.g., discounting, advantage estimation, and loss weighting) and optimization/execution (e.g., optimizer choice, batch size, and training schedule). These settings follow standard PPO practices while balancing training stability and sample efficiency.}
\begin{tabular}{l l | l l}
\toprule
\textbf{RL Strategy} & \textbf{Value} & \textbf{Optimization / Execution} & \textbf{Value} \\
\midrule
Algorithm & PPO & Optimizer & Adam \\
Discount factor $\gamma$ & 0.99 & Learning rate & $5 \times 10^{-4}$ \\
GAE $\lambda$ & 0.95 & Mini-batch size & 256 \\
Entropy coefficient & 0.01 & PPO epochs & 10 \\
Value loss coefficient & 0.5 & Grad clip (norm) & 0.2 \\
Episode length & 2048 & Training time & 24 hr \\
\bottomrule
\end{tabular}
\label{tab:training_config_compact}
\end{table}

\section{Training Configuration}
\label{sec:train_details}

All reinforcement learning experiments are conducted using Proximal Policy Optimization (PPO), chosen for its stability and efficiency in on-policy policy gradient settings. Table~\ref{tab:training_config_compact} summarizes the training configuration adopted uniformly across all tasks. The hyperparameters are grouped into two functional categories: RL strategy and optimization/runtime settings.

In the RL strategy group, we use a discount factor of $\gamma = 0.99$ and generalized advantage estimation (GAE) with $\lambda = 0.95$ to stabilize learning and reduce variance. The entropy coefficient and value loss coefficient are set to 0.01 and 0.5, respectively, encouraging sufficient exploration while maintaining value accuracy.

Optimization is performed using the Adam optimizer with a learning rate of $5\times10^{-4}$. Each policy update involves 10 PPO epochs with a mini-batch size of 256 and gradient clipping at a maximum norm of 0.2 to ensure training stability. Episodes consist of 2048 environment steps, and A typical training run consists of 100,000 episodes and completes within 24 hours.

To ensure efficiency and robustness during training, all policy networks are trained on small lattice environments with dimensions $10 \times 10 \times 10$. This design choice allows rapid sampling and policy iteration without compromising the learned policy's ability to generalize. Once trained, the learned policy is directly transferred to larger physical systems (up to $5.4 \times 10^{10}$ atoms) without any additional fine-tuning, demonstrating strong scalability and transferability.

\newpage

\bibliographystyle{unsrt}  

\bibliography{base}


\newpage
\section*{NeurIPS Paper Checklist}

\begin{enumerate}

\item {\bf Claims}
    \item[] Question: Do the main claims made in the abstract and introduction accurately reflect the paper's contributions and scope?
    \item[] Answer: \answerYes{} 
    \item[] Justification: The abstract and introduction accurately capture the paper’s key innovations—including the agent-based formulation, reweighting mechanism, CTDE training, and large-scale benchmark results. All stated contributions are later substantiated with technical details and quantitative evidence.

\item {\bf Limitations}
    \item[] Question: Does the paper discuss the limitations of the work performed by the authors?
    \item[] Answer: \answerYes{} 
    \item[] Justification: Yes, please see Sec.~\ref{sec:limitations} for limitations. 

\item {\bf Theory assumptions and proofs}
    \item[] Question: For each theoretical result, does the paper provide the full set of assumptions and a complete (and correct) proof?
    \item[] Answer: \answerYes{} 
    \item[] Justification: Yes, the proof of statistical consistency is provided in Sec.~\ref{subsec:thermo}.

    \item {\bf Experimental result reproducibility}
    \item[] Question: Does the paper fully disclose all the information needed to reproduce the main experimental results of the paper to the extent that it affects the main claims and/or conclusions of the paper (regardless of whether the code and data are provided or not)?
    \item[] Answer: \answerYes{} 
    \item[] Justification: We describe the simulation environment, physical settings in Sec.~\ref{sec:experiments} and Sec.~\ref{sec:exp_details}, and the full structure of our reinforcement learning algorithm in Sec.~\ref{sec:reinforcement} and Sec.~\ref{sec:train_details}.


\item {\bf Open access to data and code}
    \item[] Question: Does the paper provide open access to the data and code, with sufficient instructions to faithfully reproduce the main experimental results, as described in supplemental material?
    \item[] Answer: \answerYes{} 
    \item[] Justification: The full source code, along with detailed instructions for reproducing all experiments and figures, will be released upon paper acceptance. We are actively preparing the repository to ensure clarity and usability for the community.

\item {\bf Experimental setting/details}
    \item[] Question: Does the paper specify all the training and test details (e.g., data splits, hyperparameters, how they were chosen, type of optimizer, etc.) necessary to understand the results?
    \item[] Answer: \answerYes{} 
    \item[] Justification: The Experimental setting is discribed in Sec.~\ref{sec:exp_details} and Sec.~\ref{sec:train_details}.

\item {\bf Experiment statistical significance}
    \item[] Question: Does the paper report error bars suitably and correctly defined or other appropriate information about the statistical significance of the experiments?
    \item[] Answer: \answerNo{} 
    \item[] Justification: Due to computational constraints, we do not report error bars. As noted in Sec.~\ref{sec:super}, each experiment involves billion-atom simulations at supercomputer scale, making repeated runs prohibitively expensive.

\item {\bf Experiments compute resources}
    \item[] Question: For each experiment, does the paper provide sufficient information on the computer resources (type of compute workers, memory, time of execution) needed to reproduce the experiments?
    \item[] Answer: \answerYes{} 
    \item[] Justification:  All experiments were conducted on a single NVIDIA A100 GPU. We report detailed memory usage (CPU and GPU) for experiment in Fig.~\ref{fig:memory_efficiency}, and specify execution settings including atom counts, vacancy concentrations, and simulation steps in Sec.~\ref{sec:experiments}. 
    
\item {\bf Code of ethics}
    \item[] Question: Does the research conducted in the paper conform, in every respect, with the NeurIPS Code of Ethics \url{https://neurips.cc/public/EthicsGuidelines}?
    \item[] Answer: \answerYes{}
    \item[] Justification: This work involves atomistic simulations using reinforcement learning and does not involve human subjects, personal data, or environmental impact. All methods, assumptions, and limitations are transparently reported in accordance with the NeurIPS Code of Ethics.

\item {\bf Broader impacts}
    \item[] Question: Does the paper discuss both potential positive societal impacts and negative societal impacts of the work performed?
    \item[] Answer: \answerYes{}
    \item[] Justification: This work facilitates efficient, high-fidelity simulation of complex materials, potentially accelerating advances in energy, aerospace, and semiconductor technologies. It lowers the barrier to large-scale modeling, making atomic-level simulation more broadly accessible.
    \begin{itemize}
        \item The answer NA means that there is no societal impact of the work performed.
        \item If the authors answer NA or No, they should explain why their work has no societal impact or why the paper does not address societal impact.
        \item Examples of negative societal impacts include potential malicious or unintended uses (e.g., disinformation, generating fake profiles, surveillance), fairness considerations (e.g., deployment of technologies that could make decisions that unfairly impact specific groups), privacy considerations, and security considerations.
        \item The conference expects that many papers will be foundational research and not tied to particular applications, let alone deployments. However, if there is a direct path to any negative applications, the authors should point it out. For example, it is legitimate to point out that an improvement in the quality of generative models could be used to generate deepfakes for disinformation. On the other hand, it is not needed to point out that a generic algorithm for optimizing neural networks could enable people to train models that generate Deepfakes faster.
        \item The authors should consider possible harms that could arise when the technology is being used as intended and functioning correctly, harms that could arise when the technology is being used as intended but gives incorrect results, and harms following from (intentional or unintentional) misuse of the technology.
        \item If there are negative societal impacts, the authors could also discuss possible mitigation strategies (e.g., gated release of models, providing defenses in addition to attacks, mechanisms for monitoring misuse, mechanisms to monitor how a system learns from feedback over time, improving the efficiency and accessibility of ML).
    \end{itemize}
    
\item {\bf Safeguards}
    \item[] Question: Does the paper describe safeguards that have been put in place for responsible release of data or models that have a high risk for misuse (e.g., pretrained language models, image generators, or scraped datasets)?
    \item[] Answer: \answerNA{}
    \item[] Justification: This work does not involve pretrained language models, generative image systems, or large-scale scraped datasets. It presents a reinforcement learning framework for atomistic simulation, which poses minimal risk of misuse.

\item {\bf Licenses for existing assets}
    \item[] Question: Are the creators or original owners of assets (e.g., code, data, models), used in the paper, properly credited and are the license and terms of use explicitly mentioned and properly respected?
    \item[] Answer: \answerYes{}
    \item[] Justification: All third-party assets used in this work—including baseline code, simulation environments, and referenced datasets—are properly cited in the paper. Their licenses and terms of use have been fully respected.

\item {\bf New assets}
    \item[] Question: Are new assets introduced in the paper well documented and is the documentation provided alongside the assets?
    \item[] Answer: \answerYes{}
    \item[] Justification: We introduce a new simulation and learning framework for large-scale atomistic modeling. While the code and assets are not yet publicly released, we plan to open-source them upon acceptance. Comprehensive documentation and usage instructions will be provided to ensure accessibility and reproducibility.
    \begin{itemize}
        \item The answer NA means that the paper does not release new assets.
        \item Researchers should communicate the details of the dataset/code/model as part of their submissions via structured templates. This includes details about training, license, limitations, etc. 
        \item The paper should discuss whether and how consent was obtained from people whose asset is used.
        \item At submission time, remember to anonymize your assets (if applicable). You can either create an anonymized URL or include an anonymized zip file.
    \end{itemize}

\item {\bf Crowdsourcing and research with human subjects}
    \item[] Question: For crowdsourcing experiments and research with human subjects, does the paper include the full text of instructions given to participants and screenshots, if applicable, as well as details about compensation (if any)? 
    \item[] Answer: \answerNA{}
    \item[] Justification: This work does not involve crowdsourcing or research with human subjects.

\item {\bf Institutional review board (IRB) approvals or equivalent for research with human subjects}
    \item[] Question: Does the paper describe potential risks incurred by study participants, whether such risks were disclosed to the subjects, and whether Institutional Review Board (IRB) approvals (or an equivalent approval/review based on the requirements of your country or institution) were obtained?
    \item[] Answer: \answerNA{}
    \item[] Justification: This work does not involve research with human subjects and therefore does not require IRB or equivalent review.

\item {\bf Declaration of LLM usage}
    \item[] Question: Does the paper describe the usage of LLMs if it is an important, original, or non-standard component of the core methods in this research? Note that if the LLM is used only for writing, editing, or formatting purposes and does not impact the core methodology, scientific rigorousness, or originality of the research, declaration is not required.
    \item[] Answer: \answerNA{}
    \item[] Justification: This work does not involve the use of large language models in the development of core methods.

\end{enumerate}
\end{document}